\documentclass{article}
\usepackage{spconf,amsmath,graphicx,booktabs,array,multirow,balance,cite,hyperref,placeins}

\title{ADAPT Semantics, NOT STRUCTURE: FEW-INSTANCE SCHEMA CALIBRATION\\FOR SCIENTIFIC PDF EXTRACTION}

\name{
Zixiao Dong$^{1,3}$,
Wei Yang$^{2,3}$,
Zihao Liu$^{1,3}$,
Chenshu Li$^{1,3}$,
Longzhang Liu$^{1,3}$,
Tao Tan$^{4}$,
Hong Xie$^{1,3*}$
\thanks{
$*$ Corresponding author: Hong Xie (hongx87@ustc.edu.cn).
}
}

\address{
$^{1}$School of Computer Science and Technology, University of Science and Technology of China\\
$^{2}$University of Science and Technology of China, 
$^{3}$State Key Laboratory of Cognitive Intelligence\\
$^{4}$CCCC Second Highway Consultants Co., Ltd.
}

\newcommand{\method}{CPSE}

\begin{document}
\maketitle

\begin{abstract}
A well-designed extraction schema is not necessarily ready for reliable LLM
execution. When only limited verified extractions are available, manually tuning
hundreds of field definitions through trial and error is costly. We
frame this problem as few-instance schema calibration: adapting the operational
semantics of an existing schema from a few annotated documents while preserving
its structural contract. We introduce \method{}, a contract-preserving semantic
extraction framework that jointly calibrates extraction prompts and field-level semantic descriptions
from a few gold annotations. \method{} decomposes the schema into an invariant
structural contract and mutable field semantics, and further separates identity
discovery from record completion using manifest-conditioned resolution. On
expert-annotated polymer-science documents, \method{} improves extraction by 9.93
points over an execution-matched baseline, with consistent gains under an
independent judge and in a blinded expert audit. These results show that \method{}
enables low-resource schema execution while preserving the output
structure required downstream.
\end{abstract}


\section{Introduction}
Scientific information extraction (IE) poses a dual challenge: designing an adequate representational schema
and reliably steering large language models (LLMs) to populate it from heterogeneous document
evidence~\cite{swain2016chemdataextractor,dagdelen2024structured}. This challenge
intensifies in scientific PDFs, where information spans across text, tables,
and figures, and field interpretation depends on experimental context
\cite{hira2024tetrahedron,zhang2023literature}. Even a well-specified schema often
resists faithful LLM execution because field-level extraction semantics—including evidence assignment, value interpretation, and entity association—are rarely specified
\cite{lin2026schema,yuksekgonul2024textgrad,shrimal2025parse}.

Reliable schema execution depends on adapting these operational semantics to
target documents, yet sufficient verified annotations are costly, particularly
for schemas with hundreds of fields. Existing approaches optimize task instructions
\cite{pryzant2023protegi,yang2023opro,yuksekgonul2024textgrad,
opsahlong2024mipro,agrawal2026gepa} or modify schema representations
\cite{shrimal2025parse}. The former improves global guidance but leaves
field-level interpretation unresolved, while the latter risks altering the
downstream interface. This gap points to a distinct post-design problem:
refining an established schema's operational semantics under limited
supervision, without altering its structure.

We refer to this problem as \emph{few-instance schema calibration}. Unlike
schema induction, which determines the representation space of a task, schema calibration starts from an established or induced structure and refines the operational semantics connecting document evidence to schema fields. We define
the \emph{structural contract} as the invariant schema components required
downstream, including field paths, types, cardinalities, requiredness, and
nesting. Field descriptions remain mutable; for example, calibration can
distinguish a polymerization temperature from a later measurement temperature without changing field paths or types.

Two recurring failure modes in polymer-science PDF extraction motivate this
formulation. Global task instructions cannot precisely specify field-level evidence
requirements, particularly when similar quantities or experimental
attributes depend on context. Separately, single-pass extraction
requires the model to discover material identities and complete records
simultaneously, making record boundaries difficult to maintain in entity-dense
documents~\cite{jain2020scirex,huang2021entity,wu2024structured}. We therefore
decouple semantic calibration from record control, jointly adapting field
semantics and prompts while introducing manifest-conditioned resolution.

We introduce \textbf{contract-preserving semantic extraction (\method)} for
few-instance schema calibration. Our main contributions are:
\begin{itemize}
    \setlength{\itemsep}{0pt}
    \setlength{\parskip}{0pt}
    \setlength{\parsep}{0pt}
    \setlength{\topsep}{2pt}
    \item We formulate few-instance schema calibration as adapting field-local
    operational semantics while preserving an explicit structural contract.
    \item Our method jointly calibrates extraction prompts and field-level semantic descriptions through PDF-aware textual feedback with the schema structure held fixed.
    \item We introduce a source-ordered material manifest to separate identity discovery from full-record generation, making record boundaries explicit during extraction.
    \item Experiments on expert-annotated polymer-science PDFs show consistent gains across internal ablations, external optimizers, independent evaluation, and a blinded expert audit.
\end{itemize}

\section{Related Work}
Scientific IE spans domain-specific pipelines for extracting entities,
measurements, and relations~\cite{swain2016chemdataextractor} and recent
LLM-based structured extraction~\cite{dagdelen2024structured}. Materials-science
studies report both the promise and domain-specific limitations of few-shot
LLMs~\cite{foppiano2024materialsllm}, while polymer-specific systems extract
material--property records from abstracts or selected full-text content
\cite{shetty2023polymer,gupta2024polymerllm}. Our formulation instead targets
full-PDF extraction into a fixed nested JSON contract and adapts the extractor
from only a few annotated documents.

Automatic prompt optimization includes instruction generation and selection,
iterative textual editing, LLM-guided search, textual-feedback optimization,
and program-level optimization
\cite{pryzant2023protegi,yang2023opro,yuksekgonul2024textgrad,
opsahlong2024mipro,agrawal2026gepa}. These methods optimize instructions,
demonstrations, or other textual program components. Schema-oriented methods
optimize mutable schema representations~\cite{shrimal2025parse}, whereas CPSE
treats the output structure as an invariant contract and jointly calibrates extraction prompts and field-level operational semantics.
Recent studies suggest that structured-output schema descriptions influence
model behavior~\cite{lin2026schema}; CPSE calibrates their field-level operational
semantics while keeping the structure fixed.

Document-level IE must associate evidence distributed across a document with
the correct entities. Prior work uses cross-sentence relation
extraction~\cite{jain2020scirex}, entity-centered template generation
\cite{huang2021entity}, and multi-stage structured extraction
\cite{wu2024structured}. \method{} first constructs a source-ordered material
manifest, then resolves small, non-overlapping identity subsets separately,
making sample boundaries explicit and limiting cross-sample mixing.

Grammar-constrained decoding enforces structural validity~\cite{geng2023grammar};
\method{} instead calibrates field semantics and evidence assignment under a
fixed structure.

\section{Method}
\begin{figure*}[t]
\centering
\includegraphics[width=\textwidth]{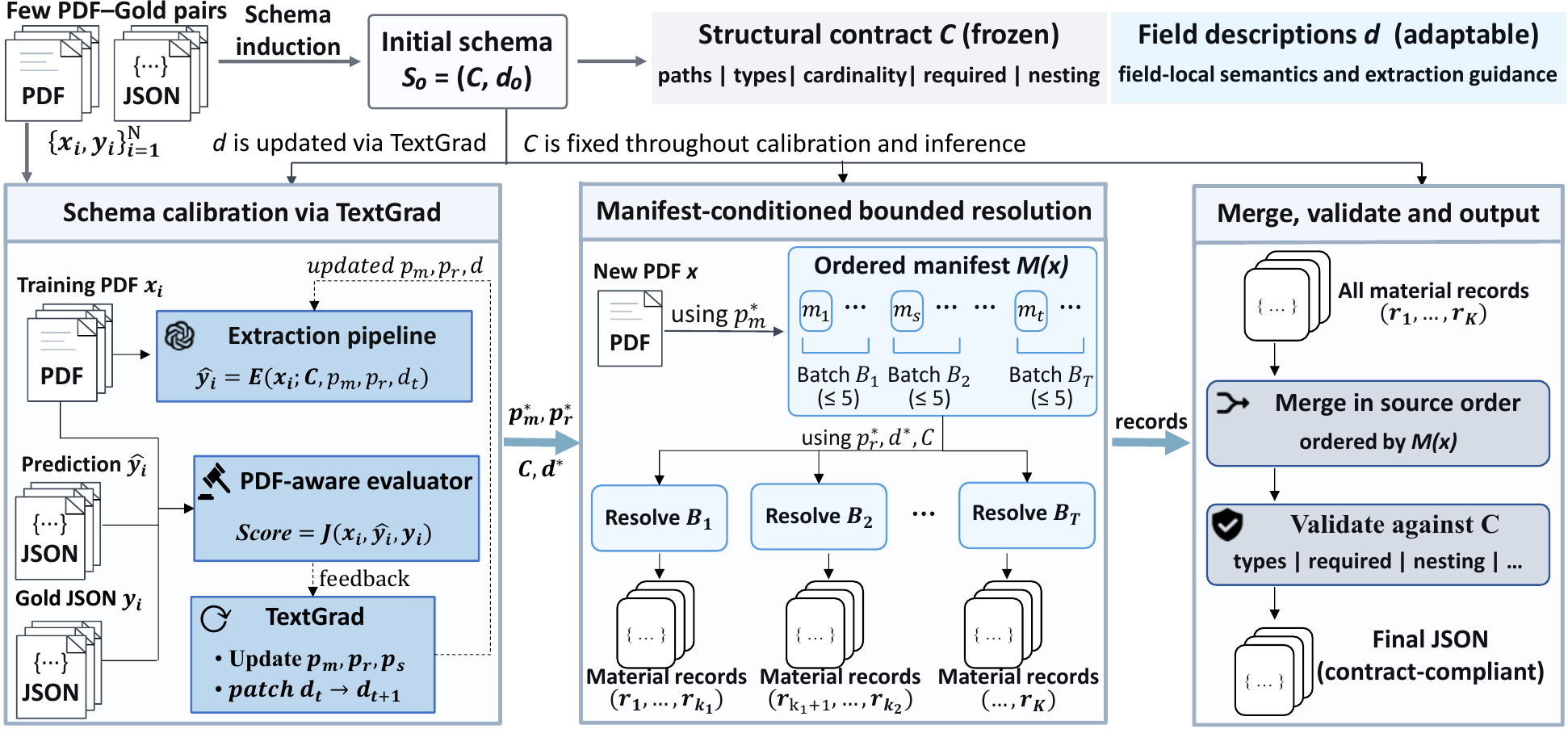}
\caption{
Overview of CPSE for few-instance schema calibration. Schema induction produces
$S_0=(C,d_0)$, after which the structural contract $C$ remains fixed while
textual feedback calibrates prompts and field descriptions. At inference, an
ordered material manifest is resolved in bounded batches before deterministic
merging and validation.
}
\label{fig:method}
\end{figure*}

\method{} begins with schema induction and then performs few-instance calibration
under a fixed structural contract. As illustrated in Fig.~\ref{fig:method}, the pipeline
consists of schema induction, constrained textual calibration, ordered identity
discovery, and manifest-conditioned bounded resolution.

\subsection{Problem formulation}
Let $D_{\mathrm{tr}}=\{(x_i,y_i)\}_{i=1}^{N}$ denote the training set, where
$x_i$ is a PDF and $y_i$ its gold JSON. During schema induction, the gold JSONs establish the
structural contract $C$, including field paths, types, cardinalities,
requiredness, and nesting, while an LLM generates the initial field descriptions
$d_0$. Together they form the initial schema $S_0=(C,d_0)$.

Calibration then freezes $C$ and treats $d$, initialized as $d_0$, as mutable. Let
$\Theta=(p_s,p_m,p_r,d)$ be the calibration state: $p_s$ updates $d$ and $p_m,p_r$
construct the manifest and resolve records, respectively. With
$S(\Theta)=(C,d)$, we optimize
\begingroup
\setlength{\abovedisplayskip}{4pt}
\setlength{\belowdisplayskip}{4pt}
\setlength{\abovedisplayshortskip}{4pt}
\setlength{\belowdisplayshortskip}{4pt}
\begin{equation}
\begin{aligned}
\Theta^*=\arg\max_{\Theta}\quad &\frac{1}{N}\sum_{i=1}^{N}
J\!\left(x_i,E(x_i;C,p_m,p_r,d),y_i\right)\\
\mathrm{s.t.}\quad &\operatorname{structure}(S(\Theta))=C,
\end{aligned}
\label{eq:objective}
\end{equation}
\endgroup
where $E$ is the extraction pipeline and $J$ a scalar PDF-aware scoring
function. Thus, prompts and descriptions may change without altering the
downstream JSON interface.

\subsection{Prompt roles and mutable schema semantics}
\method{} maintains three prompts and one adaptable description state.
The schema-patch prompt $p_s$ converts aggregated field-level feedback into sparse
description-only patches to $d$.

Given an input PDF $x$, the identity-discovery prompt $p_m$ outputs document
metadata and a source-ordered material manifest $M(x)=(m_1,\ldots,m_K)$.
Each entry $m_k$ identifies one material or sample using its name, class, form,
structural features, and source location. The record-resolution prompt $p_r$ converts
non-overlapping slices of up to five manifest entries into full records using
the PDF and current schema.

The adaptable descriptions $d$ specify field-local semantics, including
evidence boundaries, units, missing-value conventions, and sample association,
bridging the fixed schema structure and document evidence.
Description updates may refine these semantics but cannot add, delete, rename,
or relocate fields, ensuring every patch remains compliant with $C$.

The manifest establishes sample boundaries before full records are generated,
while resolving at most five identities per call limits neighboring samples and
per-call output complexity, reducing cross-sample mixing.

\subsection{Textual calibration}
We use TextGrad~\cite{yuksekgonul2024textgrad} for calibration. At round $t$, the extractor runs
on the training PDFs, and a PDF-aware evaluator returns scores and
root-cause feedback, with feedback aggregated into reusable rules rather than paper-specific facts.

TextGrad updates $p_m$, $p_r$, and $p_s$; $p_s$ then proposes a sparse
description-only patch to $d_t$. We retain the state with the highest mean
training score across rounds.

At inference, $p_m^*$ produces document metadata and $M(x)$, and duplicate
identities are merged. Disjoint manifest batches are resolved with $p_r^*,d^*,C$,
merged in source order, and validated against $C$.
\section{Experiments}
\subsection{Data and Calibration Setup}
The dataset contains 20 polymer-science papers with expert-curated JSON
annotations. We randomly select $N=3$ papers to form $D_{\mathrm{tr}}$ for schema induction and
calibration, reserving 17 for held-out evaluation. The structural contract
induced from the three training annotations contains 478 typed nodes, while
individual papers contain 2--54 material records. Sec.~\ref{sec:shot}
examines training-subset sensitivity within this three-paper pool.

We use GPT-5.6-Sol at temperature 0 for extraction, TextGrad optimization, and
training-time evaluation. Optimization starts from a one-sentence instruction
requiring contract-compliant JSON. Calibration runs for five rounds, after which
the selected state is frozen for held-out evaluation. Code is available at
\url{https://github.com/yyhlm/CPSE}.

\subsection{Evaluation and Comparison Setup}
We score all held-out predictions with two PDF-aware LLM judges, GPT-5.6-Sol
and GPT-5.6-Terra; because Sol also provides the calibration objective, Terra serves
as a held-out check for judge-specific optimization. Both use the same 100-point
rubric: identity (10), process (30), properties (50), and characterization
(10). The gold annotation defines the target content, while the PDF provides
evidence for values, conditions, and sample attribution. Equivalent
units and semantically equivalent descriptions are accepted; unsupported values
and incorrect attribution are penalized. We also audit learned prompts and
descriptions for exact reuse of training-specific strings or structural changes,
and measure the contract-validation pass rate.

For paired comparisons, we average three runs per document, compute 17
document-level score differences, and report a paired-bootstrap 95\% CI and an
exact sign-flip test. As an independent check, a blinded expert audit evaluates
eight predefined category-balanced evidence items from each of ten randomly
selected held-out papers, comparing the manifest baseline, GEPA+manifest, and
\method{} against the PDF and gold annotation.

Internal ablations compare prompt-only, description-only, and joint
prompt--description adaptation under single-pass extraction. The
manifest-conditioned baseline isolates the effect of staged execution, while
its gap to full \method{} reflects joint textual adaptation. OPRO~\cite{yang2023opro},
MIPROv2~\cite{opsahlong2024mipro}, and GEPA~\cite{agrawal2026gepa} share the
single-pass setup; as the best-performing external optimizer in this setting,
GEPA is also evaluated with \method{}'s manifest-conditioned execution. Fixed
three-shot in-context learning (ICL) and schema-free direct extraction serve as
non-optimized references.

\subsection{Main Results and Ablations}
Table~\ref{tab:main} summarizes the internal adaptation variants. Prompt-only,
description-only, and joint prompt--description adaptation improve the Sol
score by +6.81, +5.64, and +10.29 over the shared single-pass baseline,
respectively. The larger gain from joint
adaptation indicates that task-level prompts and field descriptions provide
complementary improvements.

For the end-to-end comparison, full \method{} reaches 90.95 versus 76.45 for
the unadapted single-pass baseline, a gain of +14.50, with 16 wins and one tie
across 17 papers. Against the execution-matched manifest-conditioned baseline of
81.02, \method{} gains +9.93, with 14 wins, two ties, and one regression;
the paired-bootstrap 95\% CI of [5.12, 16.11] remains entirely positive, with
an exact sign-flip $p=1.83\times10^{-4}$. These comparisons separate the
combined end-to-end benefit from the contribution of joint textual adaptation
with the execution strategy held fixed.
The broad pattern of wins and positive CI indicate that the
improvement is not driven by only a few high-gain papers.

Against the execution-matched baseline, schema-valid outputs increase from
94.1\% to 100\%. The calibration audit finds no exact reuse of 581 training-specific
strings, while the contract structure remains unchanged. Gains are concentrated in property and
process extraction, consistent with the sample--value--condition and
long-procedure errors targeted by the method.

\begin{table}[htbp]
\centering
\caption{Internal-ablation scores on 17 held-out PDFs.}
\label{tab:main}
\vspace{3pt}
\small
\renewcommand{\arraystretch}{1.08}
\begin{tabular*}{0.92\columnwidth}{@{\hspace{4pt}\extracolsep{\fill}}lcc@{\hspace{4pt}}}
\toprule
Method & Sol & Terra \\
\midrule
Single-pass baseline & 76.45 & 75.84 \\
Prompt adaptation & 83.26 & 82.44 \\
Description adaptation & 82.09 & 84.28 \\
Prompt + description & 86.74 & 86.64 \\
\midrule
Manifest-conditioned baseline & 81.02 & 83.88 \\
\textbf{Full \method{}} & \textbf{90.95} & \textbf{91.09} \\
\bottomrule
\end{tabular*}
\end{table}

The held-out Terra judge shows the same trend: against the execution-matched
baseline, full \method{} gains +7.21, with a 95\% CI of [4.35, 10.09], indicating
that the improvement is not specific to the training-time judge. In the blinded
expert audit, \method{} attains 78 fully correct facts versus
62 for the execution-matched manifest-conditioned baseline; eight of ten documents
improve and two tie, with no regressions (two-sided exact sign test, $p=0.0078$).

Table~\ref{tab:external} compares external optimizers and non-optimized
references under Sol and Terra evaluation.

Both judges rank the methods identically. Manifest-conditioned execution
raises GEPA from 82.12/83.03 to 85.52/85.50 under Sol/Terra. MIPROv2 full
exceeds its instruction-only variant by 5.23/5.42 points, while fixed three-shot
ICL lies between them and schema-free direct extraction ranks last. Among the
evaluated external baselines, GEPA+manifest performs best but remains
5.43/5.59 points below \method{} under Sol/Terra, with paired-bootstrap 95\%
CIs of [1.16, 9.85]/[1.04, 10.22] and exact sign-flip
$p=.0339/.0354$.
On the same 80 audited facts, GEPA+manifest attains 67 fully correct facts,
between the manifest baseline (62) and \method{} (78), matching the automatic
ranking and providing further expert support for \method{}'s advantage.

\begin{table}[htbp]
\centering
\caption{Mean Sol and Terra scores for external optimizers and non-optimized
references.}
\label{tab:external}
\vspace{3pt}
\small
\renewcommand{\arraystretch}{1.08}
\begin{tabular*}{0.92\columnwidth}{@{\hspace{4pt}\extracolsep{\fill}}lcc@{\hspace{4pt}}}
\toprule
Method & Sol & Terra \\
\midrule
GEPA prompt only~\cite{agrawal2026gepa} & 82.12 & 83.03 \\
GEPA~\cite{agrawal2026gepa} + manifest & 85.52 & 85.50 \\
MIPROv2 full~\cite{opsahlong2024mipro} & 79.44 & 77.86 \\
MIPROv2 instruction only~\cite{opsahlong2024mipro} & 74.21 & 72.44 \\
OPRO~\cite{yang2023opro} & 71.35 & 71.24 \\
Fixed 3-shot ICL & 76.50 & 74.37 \\
Schema-free direct extraction & 64.76 & 63.20 \\
\midrule
\textbf{\method{}} & \textbf{90.95} & \textbf{91.09} \\
\bottomrule
\end{tabular*}
\end{table}

\subsection{Training-Subset Sensitivity and Mechanism Analysis}
\label{sec:shot}
Against a matched manifest baseline for each subset, the three one-paper subsets
average a held-out Sol gain of +3.48, the three two-paper subsets +2.23, and all
three papers +9.93. The non-monotonic pattern indicates training-subset sensitivity, 
suggesting that factors beyond example count, potentially including subset composition
and optimization variability, affect calibration performance in this low-resource regime.

Case analyses further align with the two failure modes motivating \method{}. In
an entity-dense polyimide case, recovering record boundaries coincides with the
largest gain (+49.2), while correcting sample--condition attribution in another
case yields +20.0. These cases are consistent with improved separation and
association of closely related records, while also highlighting identity recall as a
key bottleneck: omissions in the manifest cannot be recovered downstream.
Finally, in a near-tie case, both
outputs are correct on all eight audited facts despite the evaluator difference,
indicating residual judge sensitivity when predictions are otherwise comparable.

\section{Conclusion}
We introduced \method{} for few-instance schema calibration in scientific PDF
extraction. \method{} decomposes the schema into a fixed structural contract and
mutable field semantics, then jointly calibrates prompts and field descriptions
while explicitly controlling record boundaries through manifest-conditioned
resolution. On expert-annotated polymer-science PDFs, \method{} improves
extraction across automatic and expert evaluation without changing the JSON
interface. These results show that a few verified examples can improve
schema execution without altering the downstream output contract, positioning
schema calibration as a distinct post-design stage for reliable LLM extraction.

\vfill\pagebreak
\clearpage
\noindent\textbf{Compliance with Ethical Standards.}
This study uses publicly available scientific literature for which no ethical approval was required.

\noindent\textbf{Acknowledgments.}
This work was supported by the Strategic Priority ResearchProgram of the
Chinese Academy of Sciences (Grant No. XDA0490000).
The authors declare no conflicts of interest.
\bibliographystyle{IEEEbib}
\bibliography{refs}
\end{document}